\documentclass[11pt]{article}

\usepackage[final]{acl}

\usepackage{times}
\usepackage{latexsym}

\usepackage[T1]{fontenc}

\usepackage[utf8]{inputenc}

\usepackage{microtype}

\usepackage{inconsolata}

\usepackage{graphicx}

\usepackage{amsmath}
\usepackage{booktabs} 
\usepackage{multirow} 
\usepackage{graphicx}

\usepackage[utf8]{inputenc}
\usepackage{amssymb} 
\usepackage{geometry}
\usepackage{authblk}

\makeatletter

\renewcommand*{\@fnsymbol}[1]{\ensuremath{\ifcase#1\or * \or \diamondsuit \or \heartsuit \or \clubsuit \or \spadesuit \else\@ctrerr\fi}}
\makeatother

\title{
    \texorpdfstring{
        \protect\raisebox{-0.3\height}{
            \includegraphics[height=2em]{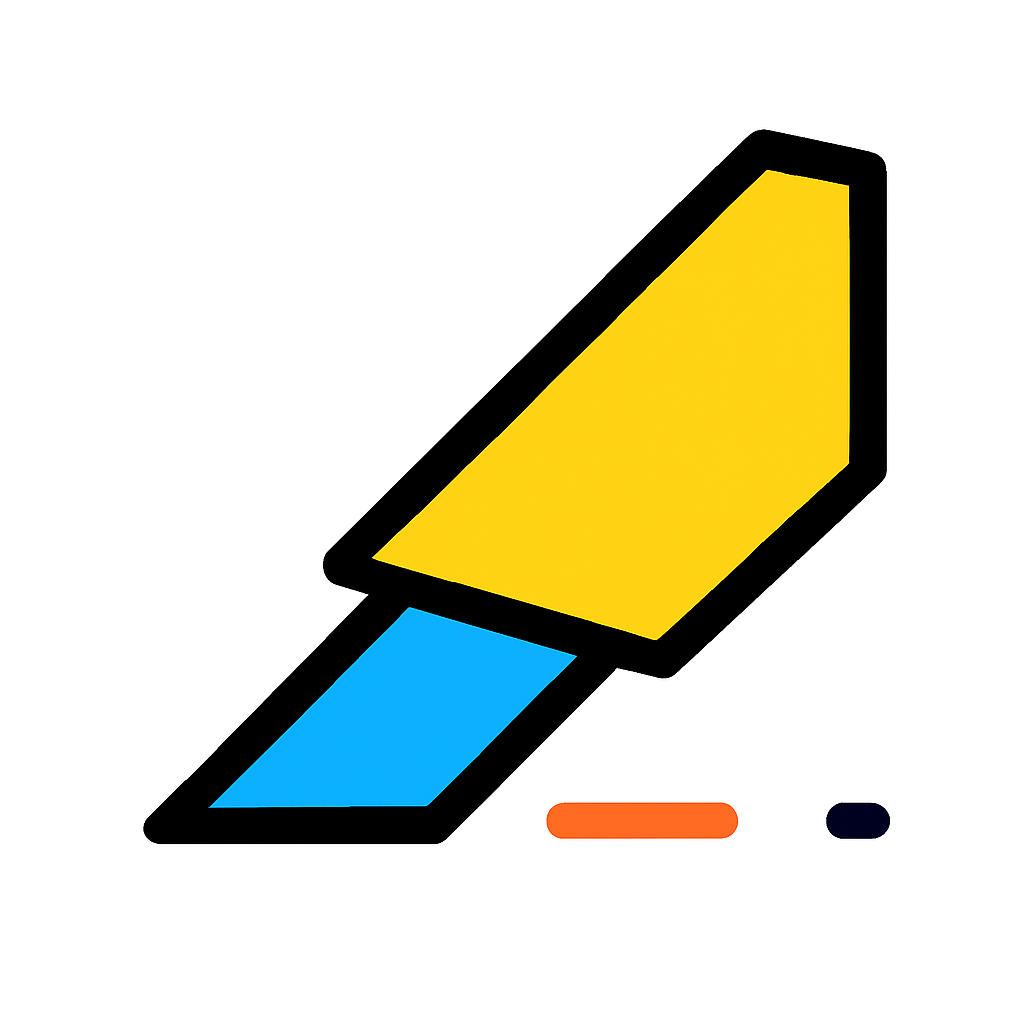}
        }
        \hspace{0em} 
    }{}
    DeepResearch-Slice: Bridging the Retrieval-Utilization Gap \\ 
    via Explicit Text Slicing\thanks{Ongoing work.} 
}

\author{
  Shuo Lu$^{\diamondsuit}$ \quad
  Yinuo Xu$^{\diamondsuit,\heartsuit}$ \quad
  Jianjie Cheng$^{\clubsuit}$ \quad
  Lingxiao He$^{\clubsuit}$ \quad
  Meng Wang$^{\clubsuit}$ \quad 
  Jian Liang$^{\diamondsuit,\heartsuit}$ \\
  \vspace{2mm}
  $^{\diamondsuit}$NLPR \& MAIS, CASIA \quad
  $^{\heartsuit}$School of AI, UCAS \quad
  $^{\clubsuit}$Meituan Inc. \\
  \vspace{2mm}
  \ttfamily\small
  \{shuolucs, YNfloxxxt, liangjian92\}@gmail.com\\
  \ttfamily\small
  chengjianjie@meituan.com, \{xiaomingzhidao1, wangmeng0627\}@gmail.com
}

\begin{document}
\maketitle

\vspace{2cm}

\begin{abstract}
Deep Research agents predominantly optimize \textit{Search Policies} to maximize retrieval probability ($P(\text{Retrieved})$). However, we identify a critical bottleneck: the \textbf{Retrieval--Utilization Gap}, where models fail to utilize gold evidence ($P(\text{Utilization} \mid \text{Retrieved})$) due to ``context blindness'' in noisy environments. To bridge this gap, we propose \textsc{DeepResearch-Slice}, a simple yet effective neuro-symbolic framework. Unlike implicit attention, our approach predicts precise span indices to execute a deterministic ``hard filter'' before reasoning. Extensive evaluations across six benchmarks show significant robustness gains. Applying our method to frozen backbones yields a $+73\%$ relative improvement ($19.1\% \to 33.0\%$), effectively mitigating noise without requiring parameter updates to the reasoning model. These results underscore the necessity of explicit grounding mechanisms in open-ended research.
\end{abstract}
\section{Introduction}
\label{sec:intro}

The rapid advancement of Large Language Models (LLMs) has precipitated a paradigm shift from simple information retrieval to \textbf{Deep Research}—a domain defined by the systematic automation of complex knowledge work through intelligent knowledge discovery, end-to-end workflow automation, and collaborative intelligence \cite{xu2025comprehensive}. Unlike traditional search engines that return a list of links, Deep Research agents (e.g., OpenAI Deep Research~\citep{openaideepresearch}, Perplexity~\citep{perplexitydeepresearch}) are designed to autonomously navigate vast information landscapes, synthesize heterogeneous data sources, and conduct multi-step reasoning to solve open-ended problems. Consequently, the prevailing research direction has focused heavily on optimizing the \textit{Search Policy}—teaching agents ``what, when, and how to search''—to maximize the acquisition of relevant information and ensure broad coverage of the topic.

\begin{figure*}
    \centering
    \includegraphics[width=0.9\linewidth]{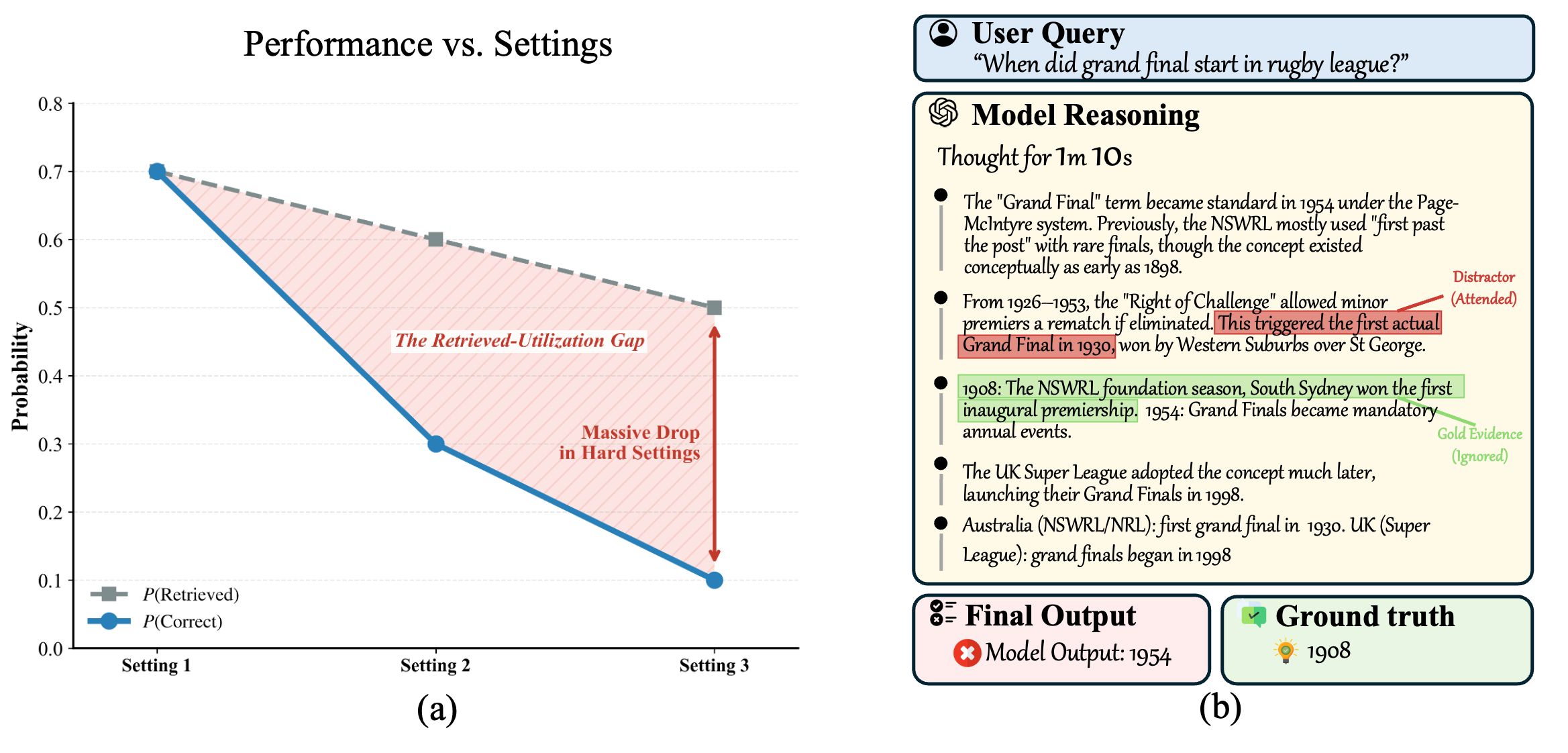}
    \caption{The "Retrieved-Utilization Gap" in Deep Research.(a) As task difficulty increases, the gap between retrieval success and final correctness expands, marking utilization as the critical bottleneck in DeepResearch. (b) A demonstration where GPT-5 succumbs to this gap, retrieving the correct document but failing to utilize it due to distraction.}
    \label{fig:figure1}
\end{figure*}

Existing approaches, ranging from standard Retrieval-Augmented Generation (RAG) pipelines~\citep{lewis2020retrieval} to autonomous agents interacting with live web search APIs, operate under a shared assumption: if the model successfully retrieves the document containing the answer, the reasoning engine will naturally derive the correct conclusion. Whether relying on vector databases or real-time Bing/Google searches \cite{xu2025comprehensive}, state-of-the-art systems prioritize expanding the search space and refining query formulation strategies~\citep{asai2024self,ma2023query} to ensure the ``Gold Evidence'' is physically present within the context window.

In this work, we revisit this assumption from a probabilistic perspective.
Let $R$ be the event that the retrieval module has fetched at least one passage containing sufficient evidence to answer the query, and $U$ be the event that the reasoning module successfully \emph{utilizes} that evidence when producing its final answer.
The overall task accuracy can then be decomposed as: $P(\text{Correct}) = P(\text{Retrieved}) \times P(\text{Utilization} \mid \text{Retrieved})$.
Existing Deep Research work is almost entirely devoted to improving $P(Retrieved)$---better search policies, better retrievers, better query reformulation.
However, our pilot studies (Section~\ref{sec:gap}) show that even the state-of-the-art (SOTA) models, including frontier LLMs such as GPT-5 and strong open-source baselines, often fail on examples where $Retrieved$ already holds: the answer is literally present in the retrieved context, yet the model either ignores it due to ``lost-in-the-middle'' effects~\citep{liu2024lost}, is distracted by spurious passages~\citep{yoran2023making,chen2022rich}, or over-trusts its parametric knowledge.
We refer to this systematic failure of models to exploit already-available evidence as the \textbf{Retrieval--Utilization Gap}.
Figure~\ref{fig:figure1} visualizes this decomposition and highlights concrete failure cases where agents ``see but do not use'' the correct document~\footnote{The specific difficulty setting will be elaborated later.}.

To explicitly target this gap, we propose \textbf{DeepResearch-Slice}, a simple but effective framework that focuses on improving $P(\text{Utilization} \mid \text{Retrieved})$ while remaining largely agnostic to how retrieval is performed.
Instead of asking the model to attend over an entire, noisy context, we introduce an explicit \emph{text slicing} mechanism: the agent first predicts a small set of token-span indices that are most likely to contain the answer, and a deterministic parser then extracts only these slices from the retrieved documents.
The reasoning model receives this sliced evidence---together with the original question and intermediate thoughts---and produces the final answer.
By turning the vague notion of ``paying attention to the right evidence'' into a discrete prediction problem over indices, DeepResearch-Slice transforms utilization into a controllable, inspectable component~\citep{lei2016rationalizing} that is less vulnerable to contextual dilution.

Our contributions are three-fold:
\begin{itemize}
    \item We formalize Deep Research accuracy as $P(\text{Retrieved}) \times P(\text{Utilization} \mid \text{Retrieved})$ and empirically demonstrate that the Retrieval--Utilization Gap is a major bottleneck for state-of-the-art Deep Research agents, even when retrieval already contains gold evidence.
    \item We introduce DeepResearch-Slice, an explicit index-based slicing mechanism that decouples evidence \emph{selection} from evidence \emph{reasoning}, enabling deterministic grounding while remaining compatible with existing Deep Research architectures.
    \item We show that DeepResearch-Slice consistently improves robustness across noise levels and search settings, yielding substantial gains over strong baselinesr.
\end{itemize}

\section{Related Work}
\label{sec:related_work}

\subsection{LLM-based DeepResearch Agents}

DeepResearch agents focus on automating complex information seeking. Early frameworks established paradigms for tool usage and iterative refinement, ranging from tool-augmented chain-of-thought~\citep{nakano2021webgpt, schick2023toolformer, qin2023toolllm} and browser-based assistants~\citep{deng2023mind2web, gur2023real} to self-reflective architectures like \textsc{ReAct} and \textsc{Reflexion}~\citep{yao2022react, shinn2023reflexion}. 
Recent systems explicitly model the search process: \textsc{STORM}~\citep{shao2024assisting} synthesizes long-form articles via multi-perspective questioning, while \textsc{Googling}~\citep{gunn2018googling} and \textsc{Self-RAG}~\citep{asai2024self} interleave retrieval tokens with generation.
Furthermore, policy optimization methods (e.g., DeepResearcher~\citep{zheng2025deepresearcher}, Search-R1~\citep{jin2025search}) employ reinforcement learning or curriculum design to decide \emph{when} and \emph{what} to search~\citep{zheng2025deepresearcher, jin2025search, sun2025zerosearch}.
However, these methods typically ingest retrieved pages as monolithic contexts, relying on implicit attention. This leaves them susceptible to ``lost-in-the-middle'' effects~\citep{liu2024lost} and attention dilution~\citep{levy2024same}. 
In contrast, we target \emph{utilization}. We introduce \textsc{Slice}, a model-agnostic operator that extracts precise, index-addressable spans—akin to extractive rationale selection~\citep{lei2016rationalizing, yu2022generate} but applied to open-ended web traces. By replacing passive reading with active extraction, we mitigate the Retrieval--Utilization Gap persisting even under optimal search policies.

\subsection{Retrieval-Augmented Generation and Robustness}

Retrieval-Augmented Generation (RAG) enhances LLM factuality~\citep{gao2023retrieval, lewis2020retrieval, guu2020retrieval, izacard2023atlas} but remains sensitive to input noise. Studies show that irrelevant or contradictory passages frequently degrade reasoning, sometimes below closed-book baselines~\citep{asai2024self, hu2024rag, yoran2023making}. 
This is exacerbated in open-ended web research, where heterogeneous content (e.g., code, forums) yields lower signal-to-noise ratios than curated corpora~\citep{karpukhin2020dense}.
Existing robustness techniques—including \textsc{CRAG}~\citep{yan2024corrective}, query rewriting~\citep{ma2023query}, and reranking—operate primarily at the coarse \emph{document} level. While they filter documents, they implicitly trust the LLM to handle distractions within the remaining long contexts. 
Similarly, citation-enhanced generation~\citep{gao2023enabling, menick2022teaching} improves grounding but relies on generative citations prone to hallucination.
In contrast, DeepResearch-Slice introduces a fine-grained slicing that decomposes documents into deterministic spans. By explicitly filtering noise at the \emph{sub-passage} level, we ensure only high-utility evidence is exposed to the reasoning module without modifying the underlying retriever or LLM.
\section{The Retrieval-Utilization Gap}
\label{sec:gap}

In this section, we formalize the challenge of deep research agents through a probabilistic decomposition. We present a pilot study demonstrating that the bottleneck has shifted from retrieval to utilization, and provide a theoretical toy model to explain why explicit slicing outperforms implicit attention in high-noise regimes.

\begin{figure*}
    \centering
    \includegraphics[width=0.9\linewidth]{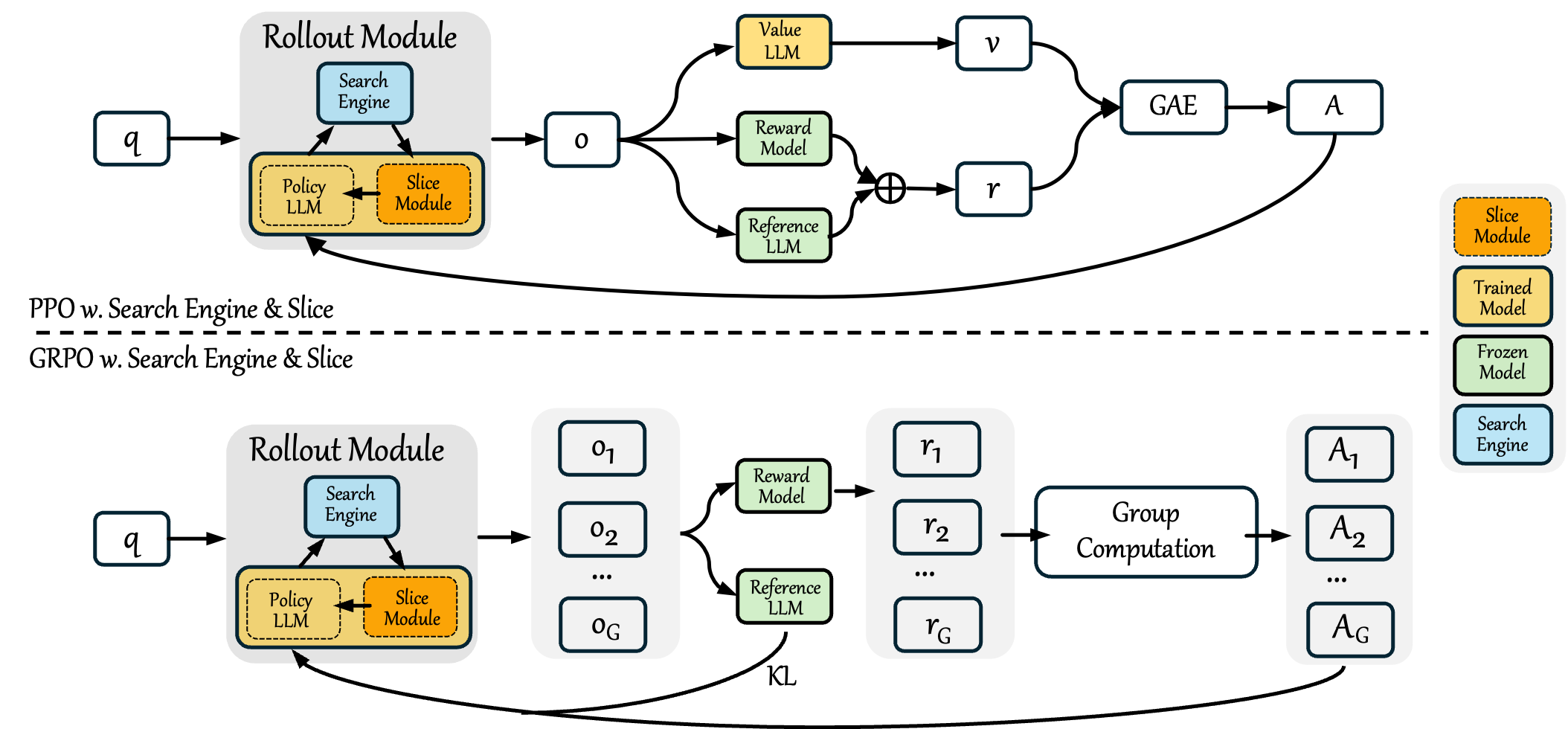}
    \caption{Demonstration of PPO and GRPO training with the search engine and slice module. The Slice Module is driven by the policy LLM shown in the diagram.}
    \label{fig:ppo_grpo}
\end{figure*}

\subsection{Decomposition Analysis}
\label{subsec:decomposition}




We formulate the probability of a correct answer, $P(\text{correct})$, as the joint probability of two sequential events: successful retrieval of the gold evidence ($R$) and the subsequent correct utilization of that evidence ($U$). This relationship decomposes as:

\begin{equation}
    P(\text{correct}) = P(R, U) = P(R) \cdot P(U \mid R).
    \label{eq:decomposition}
\end{equation}

While the prevailing paradigm in Deep Research focuses on optimizing search policies to maximize retrieval recall $P(R)$, we identify the conditional probability of utilization, $P(U \mid R)$, as the critical bottleneck in noisy environments. To validate this hypothesis, we conducted a controlled pilot study where we fixed $P(R)=1$ by injecting gold evidence into the context window amidst varying levels of distractor noise. Despite perfect retrieval, we observed a significant degradation in task accuracy as the noise ratio increased. Since $P(R)$ is constant, this performance drop is directly attributable to a decline in $P(U \mid R)$, confirming that even state-of-the-art models suffer from ``context blindness'' and ``lost-in-the-middle'' phenomena~\citep{liu2024lost} when evidence is embedded in high-noise search traces.

\subsection{Theoretical Analysis of Noise Robustness}
\label{subsec:theoretical_analysis}

To rigorously characterize the impact of slicing on information utilization, we analyze a simplified setting where the retrieved context $\mathcal{C}$ consists of $N$ segments, containing exactly one gold evidence segment $s^*$ and $N-1$ noise segments.

\paragraph{Baseline: Attention Dilution.}
Standard LLMs suffer from limited effective context windows and ``lost-in-the-middle'' phenomena~\citep{liu2024lost}. We model the implicit attention mechanism as having a finite effective capacity $T$. In high-noise regimes ($N \gg T$), the probability of the model successfully attending to $s^*$ acts as a random sampling process:
\begin{equation}
    P_{\text{base}}(\text{correct}) \approx \frac{T}{N} \cdot P(\text{correct} \mid s^*).
\end{equation}
Critically, as the volume of retrieved noise increases ($N \to \infty$), the effective recall vanishes ($\lim_{N\to\infty} \frac{T}{N} = 0$), causing performance to collapse regardless of the model's reasoning capabilities.

\paragraph{Slice: Robust Hard Filtering.}
In contrast, as shown in Figure~\ref{fig:ppo_grpo}, DeepResearch-Slice employs an explicit selection mechanism with a selection recall $q = P(\text{select } s^*)$. By isolating the evidence before reasoning, the performance lower bound is defined by:
\begin{equation}
    P_{\text{slice}}(\text{correct}) \geq q \cdot P(\text{correct} \mid s^*).
\end{equation}
It highlights a fundamental advantage: unlike the baseline, the Slice mechanism is theoretically decoupled from the total context length $N$. As long as the slicer's performance satisfies the condition $q > T/N$, our method maintains a robust performance floor even in infinite-noise environments, effectively insulating the reasoning process from retrieval distraction.

\paragraph{Conceptual Bridge}
This mechanism can be theoretically grounded as \textbf{Span-level Rationale Selection}. Prior work has shown that learning rationales over retrieved documents improves interpretability and performance. Our Slice method implements this by enforcing an index-level hard extraction rather than soft attention. This transforms the problem from ``finding a needle in a growing haystack'' (where success probability decays to 0) to a classification problem (where success probability $q$ remains stable), thereby bridging the retrieval-utilization gap.

\section{DeepResearch-Slice}

\subsection{Overview}
DeepResearch-Slice is a modular enhancement layer designed to decouple \emph{evidence retrieval} from \emph{utilization}. While existing agents ingest raw, noisy documents directly, our framework introduces an intermediate \textbf{Slice Operator} that transforms retrieved passages into precise, index-addressable spans. This process consists of three stages: (1) \textbf{Neuro-Symbolic Span Extraction}, where a model predicts start-end indices for relevant evidence; (2) \textbf{Code-Based Hard Filtering}, which deterministically parses these indices to reconstruct a clean context; and (3) \textbf{Context Integration}, where the refined evidence is fed into the reasoning model. This design is model-agnostic and plug-and-play compatible with existing search policies.

\subsection{Neuro-Symbolic Span Extraction}
\label{sec:span_extraction}

\subsubsection{Boundary Prediction as Classification}
Instead of relying on generative rewriting---which is prone to hallucination---we formulate slicing as a \emph{discrete extraction} task. Given a set of retrieved documents $\mathcal{D} = \{d_1, \dots, d_K\}$, each consisting of token sequence $\mathbf{t}^{(k)}$, we train a lightweight prediction head to identify optimal evidence boundaries.
Formally, for each document $d_k$, the model predicts a set of index triples $s_i = (k, \text{start}_i, \text{end}_i)$ that maximize the coverage of gold evidence while minimizing token usage. 
This approach constrains the action space to valid indices within retrieved documents, strictly enforcing that no new content is generated.

\subsubsection{Code-Based Hard Filtering}
To ensure zero-hallucination grounding, we employ a deterministic code parser $\textsc{Parse}(\cdot)$ to materialize the spans. The parser takes the raw documents $\mathcal{D}$ and predicted indices $\mathcal{S} = \{s_i\}_{i=1}^M$ as input, executing a verbatim extraction:
\begin{equation}
\begin{split}
  \mathcal{C}_{\text{slice}} &= \textsc{Parse}(\mathcal{D}, \mathcal{S}) \\
  &= \bigoplus_{i=1}^M \text{Tokenize}^{-1}(\mathbf{t}^{(k_i)}[\text{start}_i : \text{end}_i]).
\end{split}
\end{equation}
Unlike soft attention weights which are opaque, this operation acts as a \textbf{hard filter}, physically removing irrelevant tokens from the context window. The resulting context $\mathcal{C}_{\text{slice}}$ contains only the high-utility spans, optionally tagged with metadata (e.g., ``[Source: Doc $k$]'') to maintain traceability.

\subsection{Context Integration and Grounding}
\label{sec:integration}

The final stage integrates the sliced evidence into the reasoning workflow. Given the user query $x$, the reasoning policy $\pi_{\text{reason}}$ generates the answer $y$ conditioned on the refined context:
\begin{equation}
  y \sim \pi_{\text{reason}}( \cdot \mid x, \mathcal{C}_{\text{slice}} )
\end{equation}
This integration strategy offers three critical advantages over standard RAG pipelines:
\begin{itemize}
    \item \textbf{Noise Robustness:} By aggressively filtering distractors ($N \to 0$ for noise segments), we artificially boost the signal-to-noise ratio, mitigating ``lost-in-the-middle'' effects.
    \item \textbf{Traceable Grounding:} Every sentence in $\mathcal{C}_{\text{slice}}$ is mapped to a specific source index, enabling precise citation generation and error analysis.
    \item \textbf{Efficiency:} The reduced context length allows the reasoning model to allocate its limited attention budget solely to relevant information, improving $P(U \mid R)$ without requiring retraining of the underlying LLM.
\end{itemize}

\begin{table*}[t]
\centering
\small
\setlength{\tabcolsep}{5pt}
\begin{tabular}{lccccccc}
\toprule
\toprule
\multirow{2}{*}{\textbf{Model}} & \multicolumn{3}{c}{\textbf{Single-Hop QA}} & \multicolumn{3}{c}{\textbf{Multi-Hop QA}} & \multirow{2}{*}{\textbf{Avg}} \\
\cmidrule(lr){2-4} \cmidrule(lr){5-7}
 & \textbf{NQ$^+$} & \textbf{TriviaQA} & \textbf{PopQA} & \textbf{HotpotQA$^+$} & \textbf{2Wiki} & \textbf{Musique} & \\
\midrule
\multicolumn{8}{l}{\textit{\textbf{Inference-Only Settings}}} \\
\quad Qwen2.5-7B-Instruct & 0.134 & 0.408 & 0.140 & 0.183 & 0.250 & 0.031 & 0.191 \\
\quad \textbf{Slice-Base (Ours)} & \textbf{0.328} & \textbf{0.575} & \textbf{0.368} & \textbf{0.306} & \textbf{0.282} & \textbf{0.120} & \textbf{0.330} \\
\bottomrule
\bottomrule
\end{tabular}
\caption{Main results under the \textbf{Clean} retrieval setting for Inference-Only models. Datasets marked with $^+$ denote training sets. \textbf{Slice-Base} achieves superior performance compared to the instruction-tuned baseline.}
\label{tab:inference-results-clean}
\end{table*}

\section{Experiments}
\label{sec:exp}

\subsection{Setup}
\label{sec:setup}

\paragraph{Baselines and Architecture.}
Our primary backbone is \textbf{Qwen2.5-7B-Instruct}. We compare our method against the standard inference pipeline of this model (labeled as Qwen2.5-7B-Instruct in Table~\ref{tab:inference-results-clean}). The baseline follows a standard ``search-then-read'' approach, where full retrieved documents are directly concatenated into the context window. For our method, we introduce the \textbf{Slice-Base} module to filter the context before the reasoning step.

\paragraph{Method Configuration.}
We evaluate the \textbf{Slice-Base} configuration: a slice head trained on frozen backbones using supervised oracle spans. In this setting, the reasoning model (Qwen2.5-7B) remains frozen and receives only the sliced context. This isolates the impact of context utilization, demonstrating how effectively removing noise improves performance without requiring expensive reinforcement learning or full-parameter fine-tuning of the generator.

\subsection{Main Results}
\label{sec:main_results}

Table~\ref{tab:inference-results-clean} presents the results under the \textbf{Clean} retrieval setting.

\paragraph{Inference-Only Gains.}
Applying \textbf{Slice-Base} to the frozen Qwen2.5-7B-Instruct backbone yields a dramatic \textbf{$+73\%$} relative improvement ($19.1\% \to 33.0\%$ avg). 
\textbf{Slice-Base} consistently outperforms the baseline across all datasets. Notably, it achieves significant gains on Single-Hop benchmarks such as NQ ($13.4\% \to 32.8\%$) and TriviaQA ($40.8\% \to 57.5\%$). This validates that the primary bottleneck for standard instruction-tuned models in RAG settings is the abundance of noise in retrieved documents, which Slice effectively mitigates.

\subsection{Ablation Analysis}
\label{sec:ablation}


\paragraph{Token Efficiency.}
The Slice approach significantly reduces computational cost during inference. By feeding only relevant spans to the reasoning module, we achieve higher accuracy with significantly fewer input tokens compared to the standard RAG baseline. This efficiency gain confirms that performance improvements are driven by higher information density rather than simply more context.

\paragraph{Impact of Slice Quality.}
We observe a strong positive correlation between the intrinsic recall of the Slice head and downstream QA accuracy. This confirms that the Retrieval-Utilization Gap is directly bridged by the precision of the span selector.
\section{Conclusion}

In this work, we challenge the prevailing assumption in Deep Research that better search policies automatically guarantee better reasoning. Through rigorous error decomposition, we identify the \textbf{Retrieval--Utilization Gap} as a critical bottleneck, where high-quality retrieval is frequently negated by the model's inability to attend to evidence amidst noise.
To address this, we introduce \textsc{DeepResearch-Slice}, a neuro-symbolic framework that decouples evidence selection from reasoning via a discrete, index-based ``hard attention'' mechanism. This approach transforms passive information ingestion into precise, verifiable extraction. Empirically, \textsc{DeepResearch-Slice} proves highly effective, serving as a plug-and-play module that boosts the performance of frozen backbones by over $+73\%$ relative to standard instruction-tuned baselines.
Our findings advocate for a paradigm shift in Deep Research: moving beyond ``search-more'' heuristics toward ``read-better'' strategies. By explicitly optimizing $P(\text{Utilization} \mid \text{Retrieved})$, we pave the way for agents that are not only proficient at discovery but are also robust and efficient in utilization.



\bibliography{main}

\appendix

\end{document}